%% file: main.tex
\documentclass{article}

\PassOptionsToPackage{numbers, compress}{natbib}

\usepackage[preprint]{neurips_2024}

\usepackage[utf8]{inputenc} 
\usepackage[T1]{fontenc}    
\usepackage{hyperref}       
\usepackage{url}            
\usepackage{booktabs}       
\usepackage{amsfonts}       
\usepackage{nicefrac}       
\usepackage{microtype}      
\usepackage{xcolor}         

\usepackage{multirow}
\usepackage{amsmath}
\usepackage{algorithmic}
\usepackage{marvosym}

\input{mlVecMat}

\title{CityCraft: A Real Crafter for 3D City Generation}

\author{
    Jie Deng$^{1, }$\thanks{Equal contribution; \textsuperscript{\Letter}Corresponding Author; Correspondence to gaoangwang@intl.zju.edu.cn}
    \quad
    Wenhao Chai$^{2, *}$ 
    \quad
    Junsheng Huang$^{1, *}$
    \quad
    Zhonghan Zhao$^{1, *}$
    \AND
    Mingyan Gao$^{1}$
    \quad
    Qixuan Huang$^{1}$
    \quad
    Jianshu Guo$^{1}$
    \quad
    Shengyu Hao$^{1}$
    \AND
    Wenhao Hu$^{1}$
    \quad
    Jenq-Neng Hwang $^{2}$
    \quad
    Xi Li$^{1}$
    \quad
    Gaoang Wang$^{1}$ \textsuperscript{\Letter}
    \AND \vspace{-2em}
    \\
    $^{1}$Zhejiang University \quad
    $^{2}$University of Washington \\
    \url{https://github.com/djFatNerd/CityCraft}
}

\begin{document}
\maketitle
\input{tex/0_abs}

\input{tex/1_intro}
\input{tex/2_survey}

\input{tex/3_method}
\input{tex/4_exp}
\input{tex/5_conclusion}



\newpage
{\small
\bibliographystyle{plain}
\bibliography{ref}
}



\input{tex/6_appendix}


\end{document}

%% file: mlVecMat.tex
\usepackage{amsmath}
\usepackage{amsfonts}
\usepackage{amssymb}
\usepackage{wrapfig}
\usepackage{subcaption}
\usepackage{multirow}
 \usepackage{mathtools} 

\usepackage{verbatim}

\usepackage{anyfontsize}

\usepackage{microtype}
\usepackage{graphicx}
\usepackage{booktabs} 
\usepackage{multirow}
\usepackage{amsmath,amssymb}
\usepackage{booktabs}
\usepackage{caption,subcaption}

\usepackage{xcolor}
\definecolor{mygreen}{HTML}{3cb44b}
\definecolor{skyblue}{HTML}{beffff}
\definecolor{lightgreen}{HTML}{90ee90}

\usepackage{color, colortbl}

\definecolor{emerald}{rgb}{0.31, 0.78, 0.37}

\usepackage{tcolorbox}
\usepackage{enumitem}
\setitemize{itemsep=10pt,topsep=0pt,parsep=0pt,partopsep=0pt}
\usepackage{colortbl}

\usepackage{xcolor}
\definecolor{mygreen}{HTML}{3cb44b}
\colorlet{myyellow}{green!10!orange!90!}
\makeatletter

\usepackage{tikz}
\usetikzlibrary{arrows,shapes,snakes,automata,backgrounds,fit,petri}
\usepackage{adjustbox}

\newcommand{\RN}[1]{%
	\textup{\lowercase\expandafter{\it \romannumeral#1}}%
}
\usepackage{tabu}

\newcommand{\etc}[0]{\emph{etc.}}

\newcommand{\beq}{\vspace{0mm}\begin{equation}}
\newcommand{\eeq}{\vspace{0mm}\end{equation}}
\newcommand{\beqs}{\vspace{0mm}\begin{eqnarray}}
\newcommand{\eeqs}{\vspace{0mm}\end{eqnarray}}
\newcommand{\barr}{\begin{array}}
\newcommand{\earr}{\end{array}}

\usepackage{color, colortbl}
\definecolor{Gray}{gray}{0.93}

\usepackage{lipsum}

\usepackage{pifont}
\newcommand{\cmark}{\ding{51}}%
\newcommand{\xmark}{\ding{55}}%

\usepackage{makecell}

\usepackage{xcolor,amsmath}
\usepackage[linesnumbered,ruled,vlined]{algorithm2e}
\DontPrintSemicolon

\usepackage{xcolor}
\definecolor{mygreen}{HTML}{3cb44b}

\SetKwComment{Comment}{\color{green!50!black}\# }{}

\SetKwProg{Function}{def}{:}{}

\SetKwProg{For}{for}{:}{}
\SetKwProg{If}{if}{:}{}

%% file: tex/0_abs.tex
\begin{abstract}
City scene generation has gained significant attention in autonomous driving, smart city development, and traffic simulation. It helps enhance infrastructure planning and monitoring solutions.
Existing methods have employed a two-stage process involving city layout generation—typically using Variational Autoencoders (VAEs), Generative Adversarial Networks (GANs), or Transformers—followed by neural rendering. These techniques often exhibit limited diversity and noticeable artifacts in the rendered city scenes. The rendered scenes lack variety, resembling the training images, resulting in monotonous styles. Additionally, these methods lack planning capabilities, leading to less realistic generated scenes.
In this paper, we introduce \textbf{CityCraft}, an innovative framework designed to enhance both the diversity and quality of urban scene generation. Our approach integrates three key stages: initially, a diffusion transformer (DiT) model is deployed to generate diverse and controllable 2D city layouts. Subsequently, a Large Language Model (LLM) is utilized to strategically make land-use plans within these layouts based on user prompts and language guidelines. Based on the generated layout and city plan, we utilize the asset retrieval module and Blender for precise asset placement and scene construction. 
Furthermore, we contribute two new datasets to the field: 1) \textbf{CityCraft-OSM} dataset including 2D semantic layouts of urban areas, corresponding satellite images, and detailed annotations. 2) \textbf{CityCraft-Buildings} dataset, featuring thousands of diverse, high-quality 3D building assets. 
\textbf{CityCraft} achieves state-of-the-art performance in generating realistic 3D cities.
\end{abstract}

%% file: tex/1_intro.tex
\begin{figure}
  \centering
  \includegraphics[width=\textwidth]{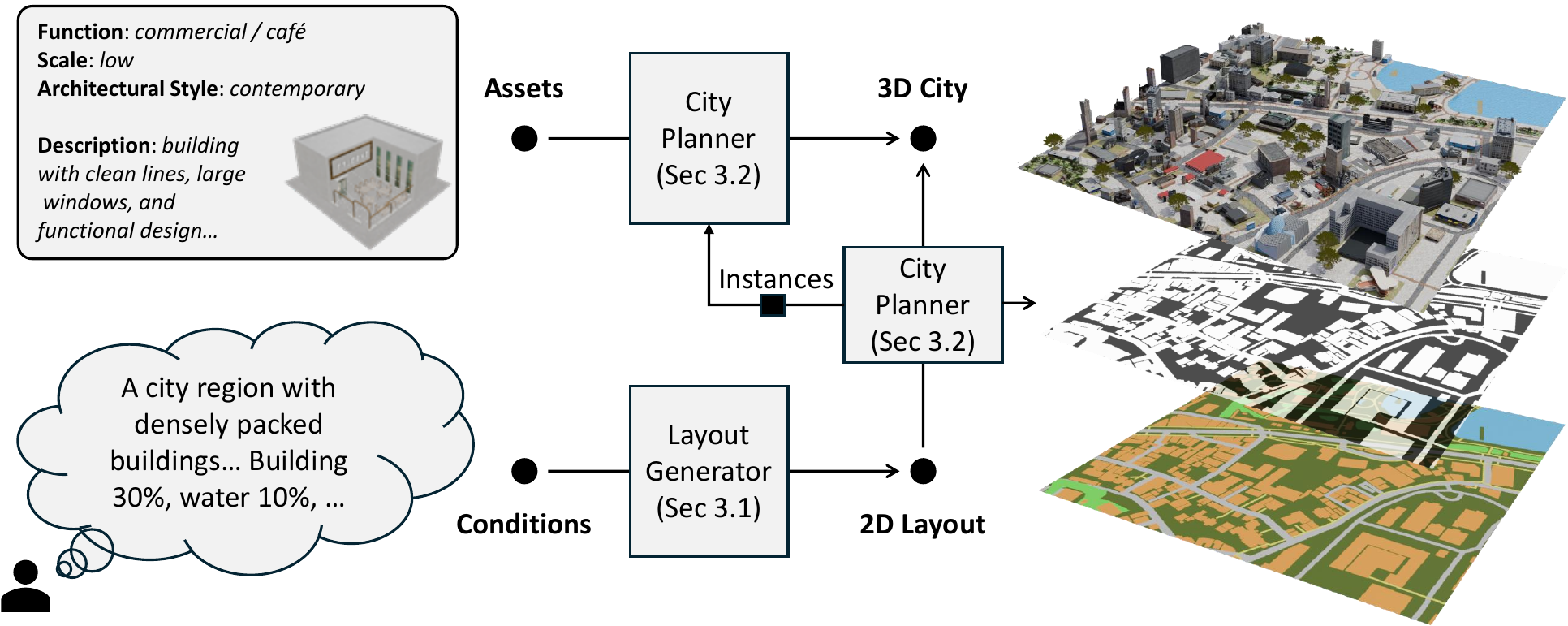}
  \caption{\textbf{Overview of CityCraft}. The \textit{Layout Generator} generates realistic 2D city layouts based on user input conditions; the \textit{CityPlanner} process the generated layouts, isolates instances, get image-level information, make land-use plans for the urban region, and select appropriate assets from assets library to craft the 3D city.}

  \label{fig:framework}
\end{figure}

\section{Introduction}

The field of 3D content generation has made remarkable progress with generative modeling, particularly in the automatic creation of 3D objects \cite{poole2023dreamfusion, ruiz2023dreambooth, Zero123}, avatars \cite{StyleaAvatar3d, DreamHuman}, and comprehensive scenes \cite{lin2023infinicity, xie2023citydreamer}. Generating 3D city scenes is crucial for simulating realistic urban environments and exploring innovative designs. Traditional methods like procedural \cite{chen2008interactive, lipp2011interactive, parish2001procedural, talton2011metropolis, yang2013urban} and image-based modeling \cite{cao2023image, aliaga2008interactive, fan2014structure, henricsson1996automated, vezhnevets2007interactive, cao2023difffashion} have established significant benchmarks, but they often lack diversity and flexibility in generating varied layout designs, hindering innovation and adaptability in urban planning.

The emerging field of text-to-3D generation combines text-to-image diffusion models with 3D representations. Innovations like DreamFusion~\cite{poole2023dreamfusion}, Magic3D~\cite{lin2023magic3d}, and ProlificDreamer~\cite{wang2023prolificdreamer} use techniques to refine 3D models and enhance the realism of generated scenes. However, applying these techniques to urban-scale scenes introduces complexities due to diverse architectural styles and complex topologies.

Recent methodologies like 3D-GPT \cite{3D-GPT} and SceneCraft~\cite{hu2024scenecraft} aim to enhance traditional approaches by combining instruction-driven 3D modeling with procedural generation software, using Large Language Models (LLMs) to assist human designers. However, challenges persist, such as the difficulty of SceneCraft \cite{hu2024scenecraft} in fine-grained editing of 3D assets and limitations of 3D-GPT in utilizing available procedural generation resources for creating large-scale, high-quality urban scenes.

Recent advancements in city scene generation~\cite{lin2023infinicity, xie2023citydreamer} have shown significant progress but highlighted the persistent challenges in creating realistic three-dimensional urban landscapes suitable for various applications. Traditional methods \cite{lin2023infinicity, xie2023citydreamer} struggle to offer enough diversity and control, leading to gaps in realism and adaptability. LLMs have the potential to revolutionize urban scene generation by providing explainable, logical, and controllable planning, enhancing the process and resulting in more adaptable, efficient, and visually appealing cityscapes.

\textbf{CityCraft} addresses these challenges by integrating state-of-the-art technologies, including Diffusion Transformers (DiT) for layout generation, Large Language Models (LLM) for strategic urban planning, and Blender~\cite{Blender} for realistic city scene crafting and rendering. This integrated approach significantly improves the generation of urban scenes that are not only diverse and detailed but also highly controllable, as shown in Figure~\ref{fig:scenes}. 

Given these advancements, the essential question we address is: \textit{How can we generate infinitely scalable, highly detailed, and controllable 3D city scenes efficiently?} In response, \textbf{CityCraft} introduces innovative methodologies to revolutionize city scene generation:

\noindent\textbf{Advanced Layout Generation:}
We use an advanced Diffusion Transformer model~\cite{peebles2023scalable} to generate high-quality, diverse and detailed 2D city layouts. The generator can take class-ratios and text as user controls and expand the generated layouts infinitely.

\noindent\textbf{Strategic Urban Planning:}
We use a Large Language Model(LLM) to implement complex city planning strategies from text prompts, resulting in more rational and realistic city generation.

\noindent\textbf{High-Quality Asset Integration and Scene Construction:}
Combining a high quality 3D assets library with Blender, CityCraft meticulously constructs city scenes, ensuring high realism and aesthetic quality through precise asset placement and advanced rendering techniques. Our main contributions are summarized as follows:

\begin{enumerate}[leftmargin=7.5mm]
\setlength{\itemsep}{2pt}
    \item We present \textbf{CityCraft}, an innovative framework that significantly enhances urban scene generation by combining advanced layout generation, strategic city planning, and high-quality scene construction.
    \item We demonstrate how \textbf{CityCraft} achieves unprecedented control and diversity in city scene generation, enabling applications requiring high detail and customization levels.
    \item We contribute two datasets, \textbf{CityCraft-OSM} and \textbf{CityCraft-Buildings}, to the community, enhancing the ability of researchers and practitioners to create more realistic and varied city environments.
\end{enumerate}

%% file: tex/2_survey.tex
\section{Related Works}
\subsection{Diffusion Models}
Denoising Diffusion Probabilistic models (DDPM)~\cite{ho2020denoising} have demonstrated exceptional success as generative models across various domains, including images~\cite{chen2023pixart, cao2023difffashion,cao2023image, chen2024pixart}, 3D objects~\cite{poole2022dreamfusion, ouyang2023chasing, vahdat2022lion}, and videos~\cite{chai2023stablevideo, luo2023videofusion}. These models have surpassed Generative Adversarial Networks (GANs)~\cite{goodfellow2014generative}, which previously dominated the field. Recent efforts in latent Diffusion Models (DMs)~\cite{rombach2022high} have shown promising applications in generating images, point clouds, and text \cite{vahdat2021score, zeng2022lion, li2022diffusionlm}. Diffusion models typically uses convolutional U-Nets~\cite{ronneberger2015u} as backbone for noise or image prediction in the denoising procedure. Recently,~\cite{peebles2023scalable} proposed to use pure transformers as the denoising network.

\subsection{Scene Planning with Large Language Models}
Recent advancements in scene design involve learning spatial knowledge priors from established 3D scene databases \cite{chang2017sceneseer,tan2019text2scene,ma2018language,tang2023diffuscene,zhao2023roomdesigner,wang2021sceneformer,wei2023lego} or refining 3D scenes iteratively based on user input \cite{chang2014interactive,cheng2019interactive}. However, approaches that rely solely on datasets like 3D-FRONT \cite{fu20213d} face limitations due to the restricted variety of categories within these datasets. 
Incorporating Large Language Models (LLMs), recent works such as LayoutGPT \cite{feng2023layoutgpt}, Holodeck~\cite{yang2024holodeck} and others \cite{lin2023towards} have demonstrated the potential of LLMs in generating 3D scene layouts and other agent-like tasks~\cite{zhao2023see,zhao2024hierarchical,zhao2024we,song2023moviechat,song2024moviechat+}. Nevertheless, direct numerical outputs from LLMs can sometimes result in physically implausible layouts, such as overlapping assets. To address this, our approach leverages LLMs to define spatial relational constraints on definite semantic layouts, utilizing a solver to optimize the arrangement. This ensures that the layouts are inspired by the vast knowledge encoded in LLMs and adhere to physical plausibility.

\subsection{City Scene Generation}
City scene generation combines detailed urban planning, including road networks, land use, and building placement, using techniques from rule-based designs \cite{bacon1976design,calthorpe2001regional} to procedural tools like CityEngine \cite{cityengine} and Unreal Engine \cite{unrealengine}, and deep learning \cite{lin2023infinicity,xie2023citydreamer,rana2023sayplan}. While diffusion models \cite{he2023diffusion,inoue2023layoutdm} often simplify layouts to basic elements, limiting complexity, Neural Radiance Fields (NeRF) \cite{lin2023infinicity, xie2023citydreamer, chen2023scenedreamer} produce high-quality visuals but are computationally expensive. In contrast, CityGen \cite{deng2023citygen} combines Stable Diffusion \cite{rombach2022high} with Low-Rank Adaptation (LoRA) \cite{hu2021lora} on the MatrixCity dataset \cite{li2023matrixcity}, resulting in more realistic and controllable city scenes through ControlNet \cite{zhang2023adding}. Additionally, new integrations of vision-language models like CLIP \cite{radford2021learning} with depth prediction \cite{fridman2023scenescape,hollein2023text2room,zhang2023text2nerf} push 3D scene generation further, though modularity remains a challenge. Our approach utilizes an extensive 3D asset database to enhance realism and interactivity, improving scene applicability for simulation, city planning, and virtual reality.

%% file: tex/3_method.tex
\section{Methods}

\subsection{City Layout Generation}

\noindent\textbf{Unconditional Generation.} In the first stage, we generate a 2D city layout for major objects in city scenes. We represent the city layout $L \in \mathbb{R}^{C \times H \times W}$ as a $H \times W$ semantic mask for $C$ classes~\cite{chen2023scenedreamer}. Considering diffusion model's excellent generation and outpainting ability, we utilize a DiT~\cite{peebles2023scalable} as the layout generator $\mathcal{D}$. We use the pretrained VAE of SDXL~\cite{podell2023sdxl} to encode training images into latent space during training and decode the denoised latents into image space during sampling. 

\noindent\textbf{Conditional Generation.} 
Condition guidance allows users to control the output, enabling the generation of images that meet user specifications. 

We introduce two types of control that can be added to our layout generator. The first is a class ratio vector $R_{L} \in \mathbb{R}^{1 \times C}$ used to enhance semantic aware generation, where $C$ is the number of classes. For a given semantic layout
$L$, the element $R_{L}[i]$ represents the ratio of class $i$, where $i \in \{1, 2, \ldots, C\}$,  $R_{L}[i] \in (0, 1)$ and $\sum_{i=1}^{C} R_{L}[i] = 1$. The second is a textual description $t_L$ that captures the desired city characteristics or specific structural elements for a given layout $L$, introduced to allow customized style generation of layouts. Directly preparing annotations for the semantic layout $L$ is challenging due to the scarcity of such data for training large vision-language models (LVMs) to generate captions. To address this issue, we utilize the corresponding satellite image $L_{sat}$ and employ GPT-4V~\cite{yang2023dawn} to generate the caption for the satellite images: $t_L= $ GPT-4V$(L_{sat})$. These conditions enhance the diffusion model's ability in reflecting user inputs, thereby improving flexibility and controllability in the generation process. The DiT blocks incorporate these conditions via adaptive layer norm~\cite{perez2018film}. The class ratio and text embeddings are mapped to the same temporal embedding space using MLP layers before being combined with the time embeddings.

\noindent\textbf{Infinite Expansion}
Our method also includes an infinite expansion feature to address the challenge of generating large-scale city layouts. We utilize BlendedDiffusion~\cite{avrahami2022blended} to achieve outpainting using the pretrained layout generation models. During sampling, we iteratively expand the city layout by generating new sections that seamlessly connect with previously generated areas, thereby enabling the creation of expansive city layouts without the limitations of fixed-size generation windows.

\subsection{City Planning}
\begin{figure}
  \centering
  \includegraphics[width=\textwidth]{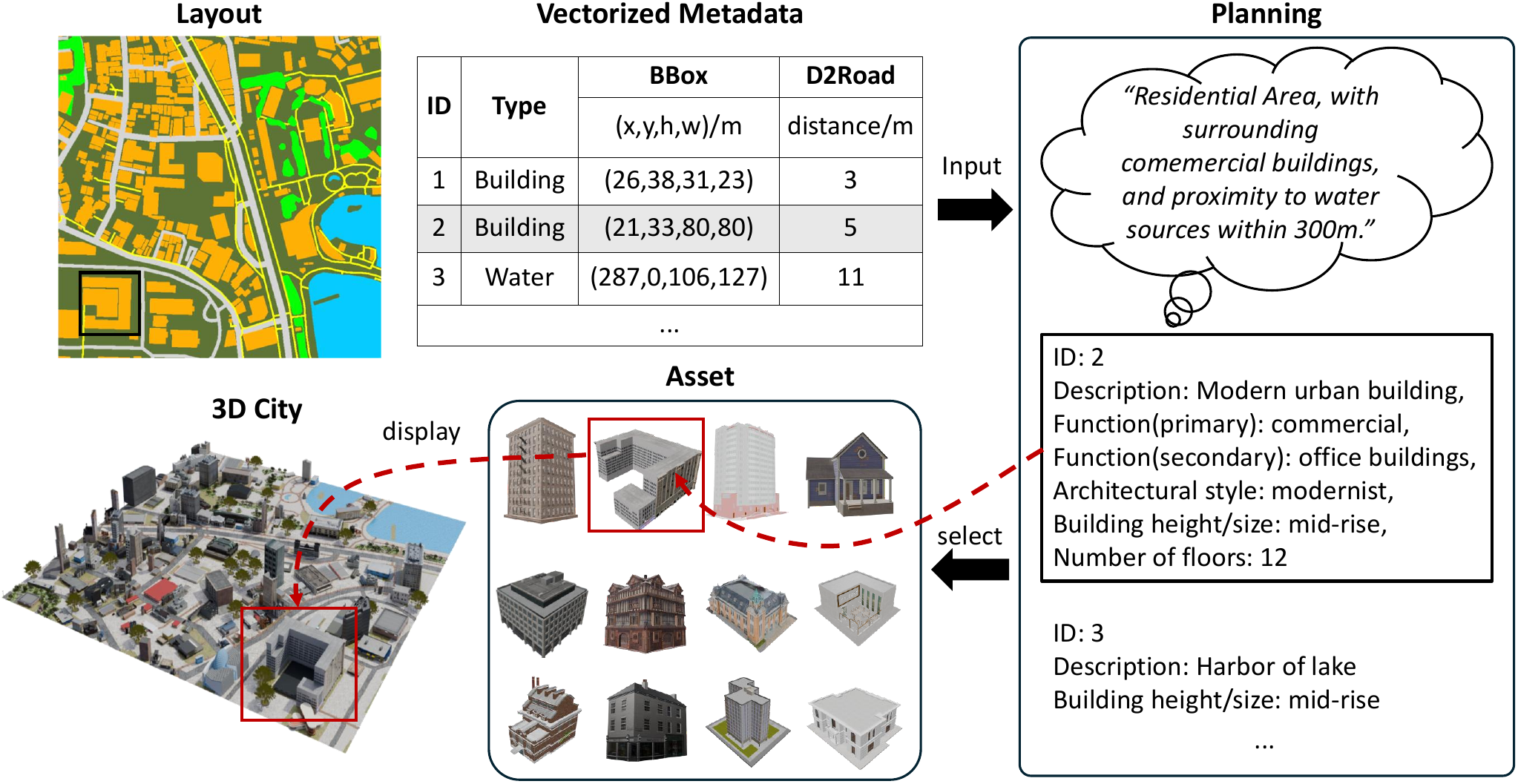}
  \caption{\textbf{Planning and Selection Process.} An example process of planning and selection. Starting from the 2D semantic layout, we isolate all instances and build a basic information dictionary for all instances based on their scale, type, spatial information, \etc (only partial information are shown in the figure for explanation, D2Road: distance to traffic roads). For each instance, we feed its information to the planner and let the planner make decisions on its characteristics. Then based on these characteristics, we retrieve the best matching candidate from the asset library and use it to craft the 3D city.}
\label{fig:planning_and_selection_process}

  \label{fig:planning}
\end{figure}

Leveraging the 2D layouts generated in the first phase, we employ an LLM as our city region planner $\mathcal{P}$ to plan the allocation of urban objects. $\mathcal{P}$ analyzes $L$'s spatial distribution and semantic context to suggest optimal space usage, which is crucial for realistic and functional urban design. Given a generated layout $L$, we first isolate all instances and retrieve information from $L$ to build a dictionary $D_L$ of information for $L$, as shown in Figure~\ref{fig:planning}. For example, for a selected instance $i$ where $i$ is a building, we extract the following information from the 2D layout: size, area, location, surrounding neighbors, distance to traffic roads, etc. We summarize this information into $D_{L}[i]$. Subsequently, we pass $D_{L}[i]$ to $\mathcal{P}$ to get plans, for instance, $i$, in this case where $i$ is a building, we output suggested planned information including primary function (residential, commercial, \etc), secondary function (store, hospital, school, \etc), size, architectural style, as well as the reasoning $\mathcal{P}$ use for making the decision. We summarize plans for instance $i$ into $P_{L}[i]$, where $P_L$ is the dictionary that stores plans for all instances. We also build an information dictionary $D_A$ for the assets library $A$, which contains image renders from multiple views, and text annotations from multiple aspects. We retrieve the most matching asset from the asset library based on the plan. During selection, we first use a tree-searching algorithm to select a subset of candidates from the asset library based on critical requirements like function and scale. Then, we calculate the similarities between these candidates and the plan. For an asset $A$ and a plan $P$, the total similarity score is calculated by Equation~\ref{eq:sim}:
\begin{equation}
    \sigma(A, P) = \sum_{m=1}^{M} w_m \cdot \sigma_{m}(\mathcal{E}_{m}^{A}, \mathcal{E}_{m}^{P})
    \label{eq:sim}
\end{equation}
where $\mathcal{E}_{m}$ is the CLIP~\cite{radford2021learning} or SBERT~\cite{reimers2019sentence} embeddings of the $m^{\text{th}}$ image or text, $w_m$ is the weight and $\sigma(\cdot)_m$ is the similarity function for pair-wised image or text embeddings. After that, the asset $A$ with the highest similarity will be selected as the candidate. 

This planning procedure is repeated several rounds for all instances, and the planner updates plans for each instance based on updated local and global information. At each iteration of the planning, we tell $\mathcal{P}$ its previous reasoning for decision and query if it wants to keep the original plan or make a new plan. This step is necessary since, in the initial steps, planner $\mathcal{P}$ only gets partial local and global information for making decisions if an instance's neighboring objects have not been planned, so the plan might not be coherent and stable in early stages. We record the number of changes made by the planner and the total number of plans at each iteration and claim convergence when their ratio is smaller than a threshold.

After making plans for all instances, we utilize Blender~\cite{Blender} to translate the 2D layout into fully realized 3D city scenes. For each chosen asset, we use Powell algorithm~\cite{powell1964efficient} to find the optimal scale and rotation factor to place it onto the 2D layout. For other objects, including vegetation, water, and roads, we add planned textures to them. We also add customized features like trees and streetlights to make the scene look more realistic.

%% file: tex/4_exp.tex
\section{Datasets}
We build two datasets to facilitate our study. The \textbf{CityCraft-OSM} dataset is used for training the city layout generation model, and the \textbf{CityCraft-Buildings} dataset is used for planning and crafting 3D cities.

\input{tab/osm.tex}

\paragraph{CityCraft-OSM dataset}
We build the CityCraft-OSM dataset upon OpenStreetMap(OSM)~\cite{OpenStreetMap}. It contains semantic layouts of real-world North America, Europe, and Asia cities. The layouts contain objects of seven classes, including terrain, vegetation, water, building, traffic road, rail, and footpath. Our dataset contains over 100,000 $768 \times 768$ patches of pixel distance 0.5 meters. We also provide semantic class ratios and the corresponding satellite images, and we annotate a subset of the satellite images using GPT4-V~\cite{openai_2023_gpt4vision}, and GeoChat~\cite{kuckreja2023geochat} with human corrections. We compare our dataset with a similar dataset CityDreamer-OSM~\cite{xie2023citydreamer}, which was also built from OSM; we show the comparison in Table~\ref{tab:osm} and qualitative comparisons in Figure~\ref{fig:osm_dataset}.

\paragraph{CityCraft-Buildings dataset}
We construct CityCraft-Buildings, consisting of 2,000 high-quality 3D building assets sourced from online open resources~\cite{Sketchfab}. This dataset encompasses buildings of diverse functions, scales, styles, \etc We provide high-quality renders for each asset from 12 angles and rich annotations detailing various features. We show examples of the assets in Figure~\ref{fig:asset_demo}.
\section{Experiments}

\subsection{Experiment Setups} We use DiT-B/2~\cite{peebles2023scalable} as the backbone for layout generation model. We use gpt-4-vision-preview~\cite{yang2023dawn} for image annotations, gpt-4o-2024-05-13~\cite{openai2023gpt4} and gpt-4-1106-preview~\cite{openai2023gpt4} for planning. All training and experiments are conducted on 8 $\times$ 4090-24GB Nvidia GPUs. The settings for all user studies and preference score calculations are described in Section~\ref{sec:user_study_settings}.

\subsection{Results on City Layout Generation}
We compare the performance of layout generation with other models that generate city layouts as semantic masks, including Infinicity~\cite{deng2023citygen}, CityDreamer~\cite{xie2023citydreamer}, and CityGen~\cite{deng2023citygen}. We do not include comparisons with other methods that use object bounding boxes to represent city layouts, such as~\cite{qu2023layoutllm, jyothi2019layoutvae, inoue2023layoutdm}. These methods typically limit object types to only buildings and roads and impose constraints on object shapes. They fail to capture intricate object shapes, diverse types, precise positions, and relationships in complex urban scenarios
, making them less suitable for city crafting.

\paragraph{Quantitative Comparison.} As shown in Table~\ref{tab:layout_fid_kid}, \textbf{CityCraft} significantly outperforms all compared methods in FID and KID scores. Specifically, \textbf{CityCraft} achieves an FID score of 27.60 and a KID score of 0.022, which are considerably lower than those of its closest competitor, CityGen~\cite{deng2023citygen}, which scores 88.38 and 0.089, respectively. This substantial improvement highlights the advanced capabilities of \textbf{CityCraft} in generating more realistic and diverse city layouts. CityCraft's preference score 8.6 further underscores its appeal, demonstrating its high user favorability. 

\paragraph{Qualitative Comparison.}
We show samples from all models in Figure~\ref{fig:layout}. InfiniCity~\cite{lin2023infinicity} lacks structured zoning, while CityDreamer~\cite{xie2023citydreamer} demonstrates enhancements with distinct grid patterns and rich semantic zones, yet it retains repetitive and simplistic features. Additionally, many buildings are interconnected, complicating instance separation, and some objects (roads, water bodies) overlap with buildings. CityGen~\cite{deng2023citygen} offers more consistency in grid layouts and a better delineation of roads but remains limited in architectural diversity. Our method, \textbf{CityCraft}, significantly enhances realism and diversity, featuring detailed, varied architectural styles and sophisticated road networks that closely mimic real city scenarios. The layouts demonstrate an advanced integration of city elements, providing a dynamic and visually appealing cityscape that outperforms other methods in aesthetics and practical layout design.


\definecolor{ground}{RGB}{85, 107, 47}
\definecolor{vegetation}{RGB}{0, 255, 0}
\definecolor{building}{RGB}{255, 165, 0}
\definecolor{rail}{RGB}{255, 0, 255}
\definecolor{trafficRoads}{RGB}{200, 200, 200}
\definecolor{footpath}{RGB}{255, 255, 0}
\definecolor{water}{RGB}{0, 191, 255}

\begin{figure}[t]
  \centering
  \includegraphics[width=\textwidth]{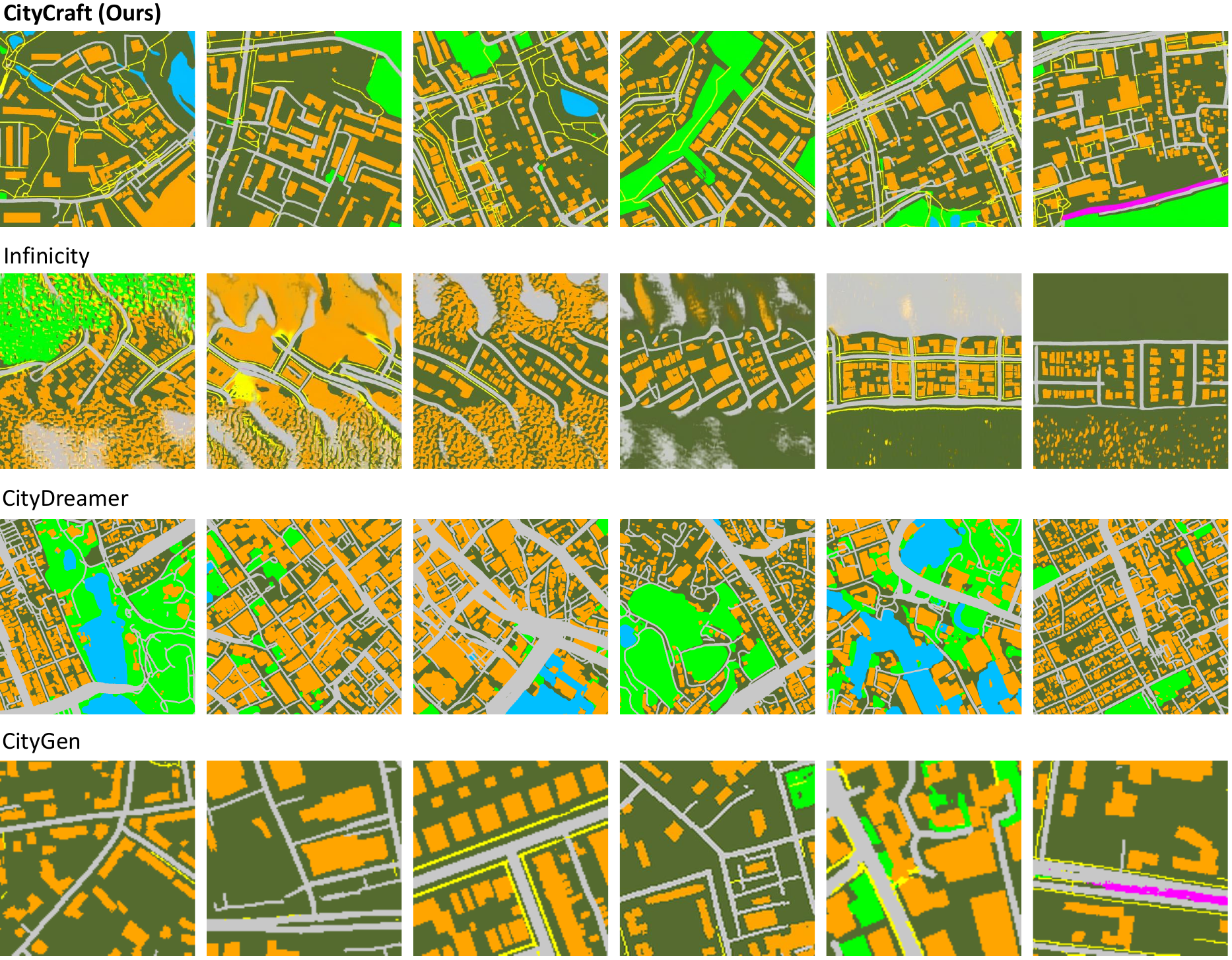}
    \parbox[c][8pt][l]{8pt}{\colorbox{ground}{}}ground\quad
    \parbox[c][8pt][l]{8pt}{\colorbox{vegetation}{}}vegetation\quad
    \parbox[c][8pt][l]{8pt}{\colorbox{building}{}}building\quad
    \parbox[c][8pt][l]{8pt}{\colorbox{rail}{}}rail\quad
    \parbox[c][8pt][l]{8pt}{\colorbox{trafficRoads}{}}traffic roads\quad
    \parbox[c][8pt][l]{8pt}{\colorbox{footpath}{}}footpath\quad
    \parbox[c][8pt][l]{8pt}{\colorbox{water}{}}water\quad
  \caption{\textbf{Qualitative comparison of city layouts.} From top to bottom: \textbf{CityCraft} (ours), InfiniCity~\cite{lin2023infinicity}, CityDreamer~\cite{xie2023citydreamer}, and CityGen~\cite{deng2023citygen}. \textbf{CityCraft} shows superior detail and realism in city planning, highlighting complex road networks and diverse architectural styles.}
  \label{fig:layout}
\end{figure}

\input{tab/layout_fid_kid}

\subsection{Results on City Scene Generation.} 
We compare the generated city scenes with other SOTA city scene generation methods, including SGAM~\cite{shen2022sgam}, Infinicity~\cite{lin2023infinicity}, CityDreamer~\cite{xie2023citydreamer}, and CityGen~\cite{deng2023citygen}.

\paragraph{Quantitative Comparison.}
We show quantitative results in Table~\ref{tab:layout_fid_kid}.
The metrics extend to Depth Error (DE)~\cite{chan2022efficient} and Camera Error (CE)~\cite{chen2023sd}, essential for assessing the accuracy of spatial and perspective representations in generated scenes. CityCraft has a Depth Error (DE) and Camera Error (CE) of \textbf{0}, showing no errors in depth or camera placement. Our framework creates precise 3D city models, enabling consistent views from any angle during sampling. In contrast, rendering-based methods such as CityDreamer~\cite{xie2023citydreamer} and Infinicity~\cite{lin2023infinicity} exhibit noticeable discrepancies and limited consistency, with the diversity of generated scenes closely resembling their training images. 
The CE and DE scores for Infinicity~\cite{lin2023infinicity} and CityGen~\cite{deng2023citygen} are not included since Infinicity~\cite{lin2023infinicity} is not open-sourced and CityGen~\cite{deng2023citygen} has no multi-view consistency as they only support synthesizing image at single view. The user preference score of 9.2 for \textbf{CityCraft} aligns with its technical metrics, further confirming user satisfaction with the realism and technical accuracy of the generated scenes.

\begin{figure}[t]
  \centering
  \includegraphics[width=\textwidth]{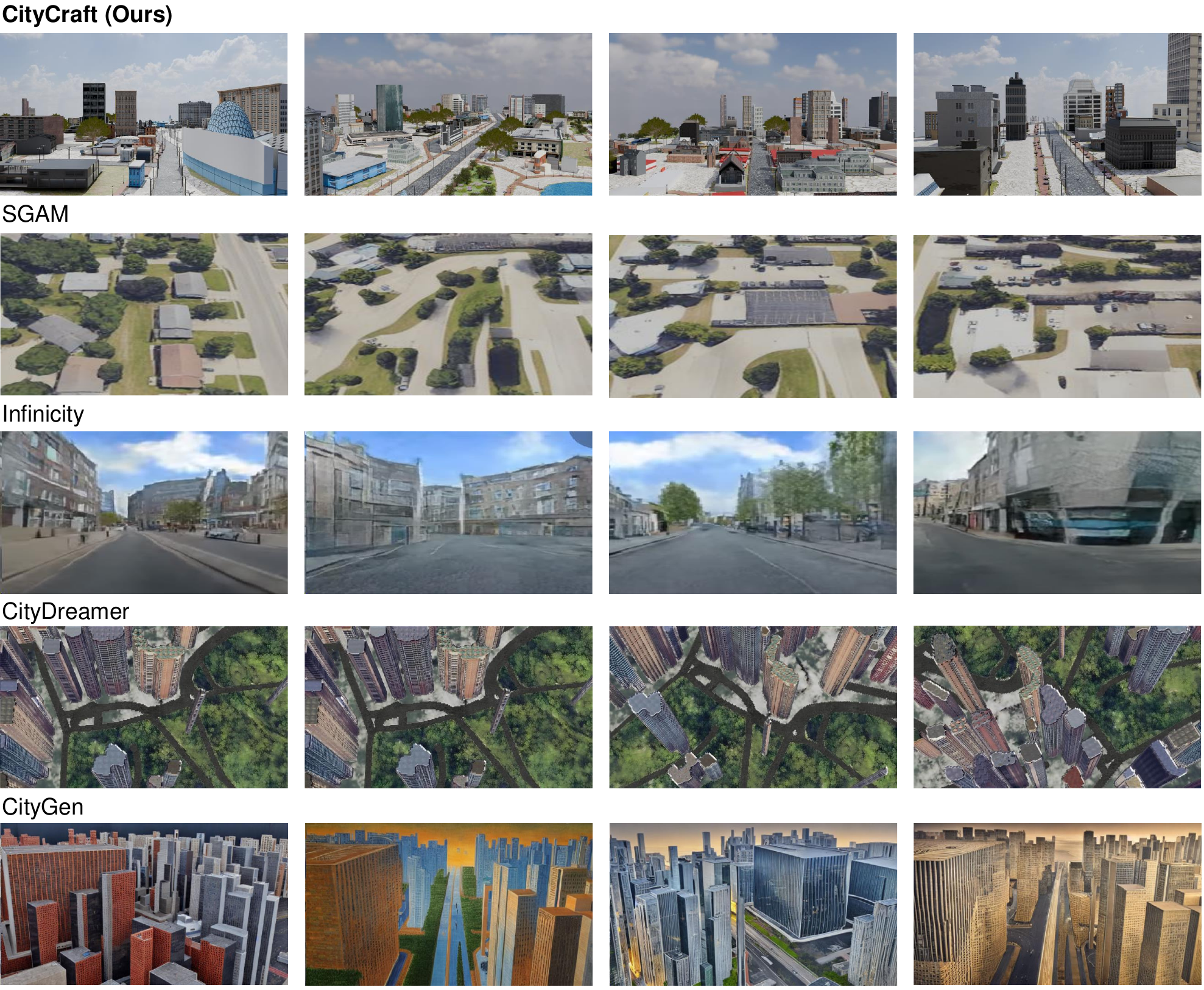}
  \caption{\textbf{Qualitative comparison of city scenes.} From top to bottom: \textbf{CityCraft} (ours), InfiniCity~\cite{lin2023infinicity}, CityDreamer~\cite{xie2023citydreamer}, CityGen~\cite{deng2023citygen}. \textbf{CityCraft} demonstrates superior architectural diversity and realism, leveraging Real 3D Crafter technology for direct building growth and LLM-driven adaptive modeling, resulting in a more authentic and varied city landscape.}
  \label{fig:scenes}
\end{figure}

\paragraph{Qualitative Comparison.}
As shown in Figure~\ref{fig:scenes}, the qualitative results underscore the superiority of \textbf{CityCraft}. Unlike InfiniCity~\cite{lin2023infinicity}, which tends to produce blurred and undetailed scenes. SGAM~\cite{shen2022sgam} and  CityDreamer~\cite{xie2023citydreamer} are much clearer. However, they still lack realism and exhibit repetitive architectural features across frames. 

\newpage
\subsection{Ablation Study}
\paragraph{Layout Generation Controls}
\input{tab/ablation_fid_kid}
We conduct an ablation study on the layout generation model under conditions, comparing the FID and KID of unconditional and conditional models by sampling 10,000 images from each. According to Table~\ref{tab:ablation_layout_fid_kid}, while FID/KID scores showed no significant differences among the models, users prefer the conditional models due to their control over layout generation. Specifically, ratio control was favored for its precise influence on the generation process. We discover that ratio-based guidance outperforms text descriptions for creating semantic layouts. We find that class ratios effectively improve layout generation and semantic understanding by calculating the Average Class Error (ACE) through the average absolute difference between generated sample class ratios and those of 10,000 input conditions.

\input{tab/ablation_control_err}
\paragraph{Scene Generation Controls}
We demonstrate the necessity of multi-round refinement for city planning. We generate 50 city scenes and measure the number of changes made by the planner in 10 rounds of planning. The results are shown in Figure~\ref{fig:trends}. From the results, we verify our assumption that the planning procedure is less stable in the initial steps and that multi-round refinement is necessary.

\begin{figure}[t]
  \centering
  \includegraphics[width=0.85\textwidth]{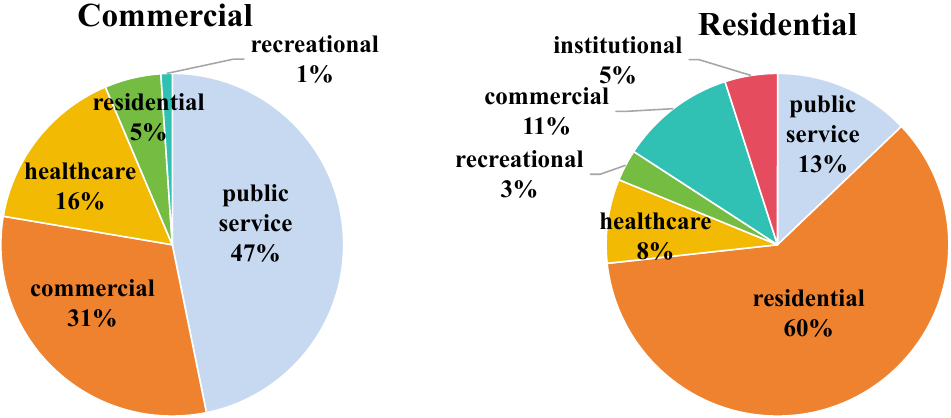}
  \caption{\textbf{City functionality distribution} with different prompts. Commercial zones are mainly for business and public services, with strong public infrastructure. Residential zones focus on living spaces, supplemented by key urban functions like healthcare and education.
}
  \label{fig:distribution}
\end{figure}

We also test the influence of global controls to generated plans.
As shown in Figure \ref{fig:distribution}, \textbf{CityCraft} demonstrates its capability to tailor urban environments through user-defined prompts. For the same city layouts, we prompt the planner to generate residential \textit{v.s.} commercial regions.
The generated commercial zones mainly comprise public service facilities (47\%) and commercial establishments (31\%), indicating a vibrant business district with concentrated economic activities. Healthcare facilities and some residential areas create a comprehensive urban environment with easy access to essential services, typical of economic hubs with strong public transport systems supporting heavy foot traffic and commerce. The generated residential zones is mainly residential (60\%), with public service facilities (13\%) and commercial spaces (11\%) for necessary amenities and leisure activities. Healthcare and educational institutions (8\% and 5\% respectively) further enhance the area's livability, reflecting a well-rounded neighborhood designed to support a thriving community. From the results, we observe the planner's ability to adapt to various user settings, proving its robustness across various scenes.

%% file: tab/osm.tex
\begin{table*}[t]
  \caption{\textbf{Quantitative comparison between CityCraft-OSM and CityDreamer-OSM~\cite{xie2023citydreamer}}. CityCraft-OSM includes a greater number of classes and a significantly larger total area. Unlike CityDreamer-OSM~\cite{xie2023citydreamer}, CityCraft-OSM also incorporates satellite images and text annotations, enhancing its utility for more detailed and accurate urban planning and analysis.}
  \resizebox{\linewidth}{!}{
    \begin{tabular}{l|cccc}
      \toprule
      Dataset  & \ Number of Classes & Total Area           & Satellite Images         & Text Annotation \\
      \midrule
      CityDreamer-OSM~\cite{xie2023citydreamer}
               & 6      & 6,000 km$^{2}$             & \xmark       & \xmark \\

      \textbf{CityCraft-OSM(Ours)} 
               & \textbf{7}      & \textbf{67,108 km$^{2}$}              & \cmark       & \cmark \\
      \bottomrule
    \end{tabular}
   }
  \label{tab:osm}
\end{table*}

%% file: tab/layout_fid_kid.tex
\begin{table*}
    \caption{\textbf{Quantitative comparison on Layout and Scene Generation}. CityGen-2 substantially outperforms other methods in layout and scene generation. Layout generation is measured by Fréchet Inception Distance(FID)~\cite{heusel2017gans} and Kernel Inception Distance(KID)~\cite{binkowski2018demystifying} (both the lower the better). DE and CE are for depth errors and camera errors adopted from~\cite{xie2023citydreamer}.}
    \centering
    \scriptsize
    \resizebox{0.8\linewidth}{!}{
    \begin{tabular}{l r r c}
        \toprule
         Method & FID $(\downarrow)$ & KID $(\downarrow)$  & Preference $(\uparrow)$ \\
         \midrule
         \textit{City Layout Generation \quad} \vspace{3pt} \\
         InfiniCity~\cite{lin2023infinicity}    & 175.68 & 0.175 & 2.3 \\
         CityDreamer~\cite{xie2023citydreamer}    & 111.44  &  0.115  & 6.8 \\
         CityGen~\cite{deng2023citygen} & 88.38 & 0.089 & 7.5\\
         \textbf{CityCraft}~(ours) & \textbf{27.60} & \textbf{0.022} & \textbf{8.6} \\
         \midrule
          Method & DE $(\downarrow)$& CE $(\downarrow)$& Preference $(\uparrow)$\\
         \midrule
         \textit{City Scene Generation \quad} \vspace{3pt}\\
         Infinicity~\cite{lin2023infinicity} & N/A & N/A & 5.1\\
         SGAM~\cite{shen2022sgam} & 0.575 & 239.291 & 6.6\\
         CityDreamer~\cite{xie2023citydreamer}& 0.147 & 0.060 & 7.6\\
         CityGen~\cite{deng2023citygen} & N/A & N/A & 5.8\\
         \textbf{CityCraft}~(ours) & \textbf{0} & \textbf{0} & \textbf{9.2}\\
         \bottomrule
    \end{tabular}
    }
    \label{tab:layout_fid_kid}
\end{table*}

%% file: tab/ablation_fid_kid.tex
\begin{table*}
    \caption{\textbf{Ablation Study of Conditional Generation}.CityCraft~(uncondition) are random samples generated from the unconditional model; CityCraft~(ratio) are conditional samples generated from user input ratios; CityCraft~(text) are conditional samples generated from user input texts.}
    \centering
    \scriptsize
    \resizebox{0.8\linewidth}{!}{
    \begin{tabular}{l c c c}
        \toprule
         Method & FID $(\downarrow)$ & KID $(\downarrow)$  & Preference $(\uparrow)$ \\
         \midrule
         CityCraft~(uncondition) & 27.60 & \textbf{0.022} & 7.9\\
         CityCraft~(ratio) & \textbf{25.58} & 0.023 & \textbf{8.7}\\
         CityCraft~(text) & 28.35 & 0.032 & 8.3 \\
         \bottomrule
    \end{tabular}
    }
    \label{tab:ablation_layout_fid_kid}
\end{table*}

%% file: tab/ablation_control_err.tex
\begin{table*}
    \caption{\textbf{Average Class Error(ACE) of ratio-controlled generation} We calculate the per average class error between the input condition class ratios and output class ratios. Results demonstrate ratios can be used as effective control for layout synthesis.}
    \centering
    \scriptsize
    \resizebox{1\linewidth}{!}{
    \begin{tabular}{l|c c c c c c c}
        \toprule
         Object Class & Ground & Vegetation & Building & Rail & Traffic Roads & Footpath & Water \\
         \midrule
          ACE(\%) & 14.94 & 5.35 & 3.78 & 5.28 & 4.48 & 11.50 & 1.78 \\
         \bottomrule
    \end{tabular}
    }
    \label{tab:ablation_control_err}
\end{table*}

%% file: tex/5_conclusion.tex
\section{Conclusion and Broader Impacts}
We introduce CityCraft, a novel framework for creating detailed 3D city scenes from user-defined text and ratio specifications. CityCraft integrates advanced techniques in layout generation, city planning, and scene construction to ensure high fidelity and alignment with user requirements. Through evaluations, it showed significant improvements over traditional methods in diversity, controllability, and visual appeal. We also introduced two novel datasets, CityGen-OSM and CityGen-Buildings, for research and development in city scene generation. We plan to expand CityCraft's capabilities to include dynamic elements such as moving traffic and integrate real-time user feedback mechanisms for interactive scene customization. 

%% file: tex/6_appendix.tex
\newpage
\renewcommand\thefigure{\Alph{section}\arabic{figure}}
\renewcommand\thetable{\Alph{section}\arabic{table}}
\setcounter{figure}{0}
\setcounter{table}{0}

\begin{center}
     \Large\textbf{Supplementary Material}
\end{center}

\noindent The supplementary material is structured as follows:

\begin{itemize}
\item We begin with the ``CityCraft Algorithm'' section, detailing the \textbf{CityCraft} method, encompassing input requirements, variables, functions, and the procedure of the \textbf{CityCraft} Method algorithm in Section~\ref{sec:pseudo}.
\item We provide ``Detailed Performance'' section for more detailed in evolution of LLM performance over iterations.
\item We deliver ``Dataset Example'' section for more detailed examples of Dataset Assets.
\item We supply ``User Study Settings'' section for details in the scoring method of preference in experiment.
\item We list ``Demo Visualization'' section for more detailed demo of city scenes generated by CityCraft.
\end{itemize}

\section{CityCraft Algorithm}~\label{sec:pseudo}

\paragraph{Preliminary}
\textbf{Diffusion Models}~\cite{ho2020denoising, rombach2022high} are state-of-the-art generative models known for their high-quality, photorealistic image synthesis. It operates on the principle of simulating a stochastic process where Gaussian noise is gradually added to an image at each step, and the model is trained to reverse this process, removing the noise and reconstructing the original image. The reverse process is typically learnt using an UNet with text conditioning support enabling text-to-image generation during inference. Specifically, given an initial noise map $\epsilon \sim \mathcal{N}(0, I)$ and a text-image pair $(c, x)$, they are trained using a squared error loss to denoise a variably-noised image as follows: 
\begin{equation}
    \mathbb{E}_{x,c,\epsilon,t}[w_t||\hat{x}_\theta(\alpha_t x + \sigma_t \epsilon, c) - x ||_2^2]
    \label{equation 1}
\end{equation}
where x is the ground-truth image, c is a text prompt, and $\alpha_t, \sigma_t, w_t$ are terms that control the noise schedule and sample quality.

\paragraph{Method}
The \textbf{CityCraft} framework is designed to generate detailed and controllable urban 3D scenes from user-defined parameters. CityCraft transforms initial textual and ratio-based inputs into complex urban environments using layout generation, urban planning, and scene construction techniques. Here, we present the pseudo-code that encapsulates the entire process.

As outlined in Algorithm~\ref{alg:citycraft}, \textbf{CityCraft} implements a three-stage framework $\mathcal{F}$ designed to convert user requirements into detailed 3D urban scenes. This transformation process is formulated as follows:
\begin{equation}
    Y = \mathcal{F}(X), \quad X \in \{\mathbf{Text}, \mathbf{Ratio}\},
\end{equation}
where $\mathbf{Text}$ represents user requirements for urban design, and $\mathbf{Ratio} \in \mathbb{R}^{1 \times C}$ denotes the probabilities of each class in a semantic layout $L$, with $C$ indicating the number of classes.

In detail, the process begins with generating semantic city $\mathbf{Layout}$, employing a layout generator $\mathcal{G}$. This generator processes an initial noise vector $\epsilon \sim \mathcal{N}(0, I)$ and a condition vector $\mathbf{Condition}$:
\begin{equation}
    \mathbf{Layout} = \mathcal{G}(\epsilon, \mathbf{Condition}), \quad \mathbf{Condition} = \mathcal{E}(X), 
\end{equation}
where $\mathbf{Condition}$ is defined as text embeddings and probabilities of each prototype class from encoding the user input $X$ with encoder $\mathcal{E}$.

After generating the initial layout, the urban planning module $\mathcal{P}$ refines $\mathbf{Layout}$ containing user requirement information to create a more detailed $\mathbf{Plan}$:
\begin{equation}
    \mathbf{Plan} = \mathcal{P}(\mathbf{Layout}),
\end{equation}

Following the planning phase $\mathbf{Plan}$, the self-adapting system $\mathcal{A}dapt$ selects and assigns appropriate assets for $\mathbf{Layout}$. This system ensures that each asset is perfectly integrated into the urban 3D scene $Y$, conforming to both user specifications and spatial dynamics:
\begin{equation}
    Y = \mathcal{A}dapt(\mathbf{Plan}, \mathbf{Layout}),
\end{equation}
where the final 3D scene $Y$ is rendered based on explainable $\mathbf{Plan}$ and original $\mathbf{Layout}$, completing the transformation from concept to visualization.

\input{pseudocode/method}

\section{Detailed Performance}

\begin{figure}[t]
  \centering
  \includegraphics[width=0.6\textwidth]{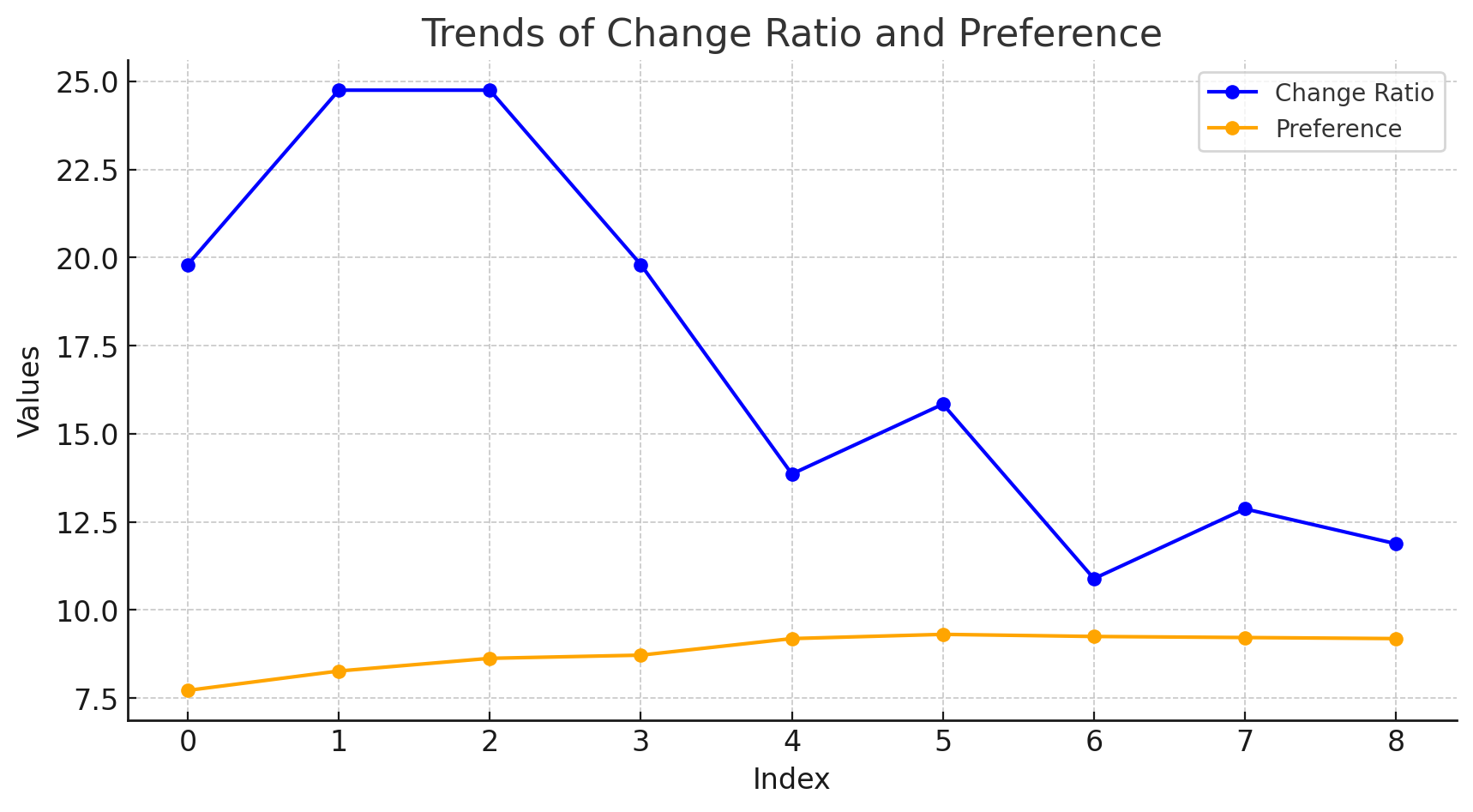}
  \caption{\textbf{Evolution of LLM performance over iterations.} The graph illustrates the relationship between the iteration index and the LLM's output dynamics. As the iteration index increases, the rate of change stabilizes, indicating a maturation in learning, while the quality of the outputs continues to rise steadily. This trend underscores the LLM's increasing efficiency and effectiveness in generating high-quality results as it undergoes more iterations.}
  \label{fig:trends}
\end{figure}

As shown in Figure~\ref{fig:trends}, one represents the rate of change, and the other illustrates the output quality as a function of the LLM's iteration index. As the index increases, indicating more iterations of the LLM, the rate of change curve likely shows a decline or stabilization, suggesting that the model is reaching a point of diminishing returns in terms of learning new patterns or making significant adjustments to its outputs. Meanwhile, the quality curve is expected to show a steady increase, indicating that the overall performance and output quality of the LLM are improving with each iteration despite the slower rate of change.

\section{Dataset Example}
\label{sec:dataset_details}

\begin{figure}
  \centering
  \includegraphics[width=\textwidth]{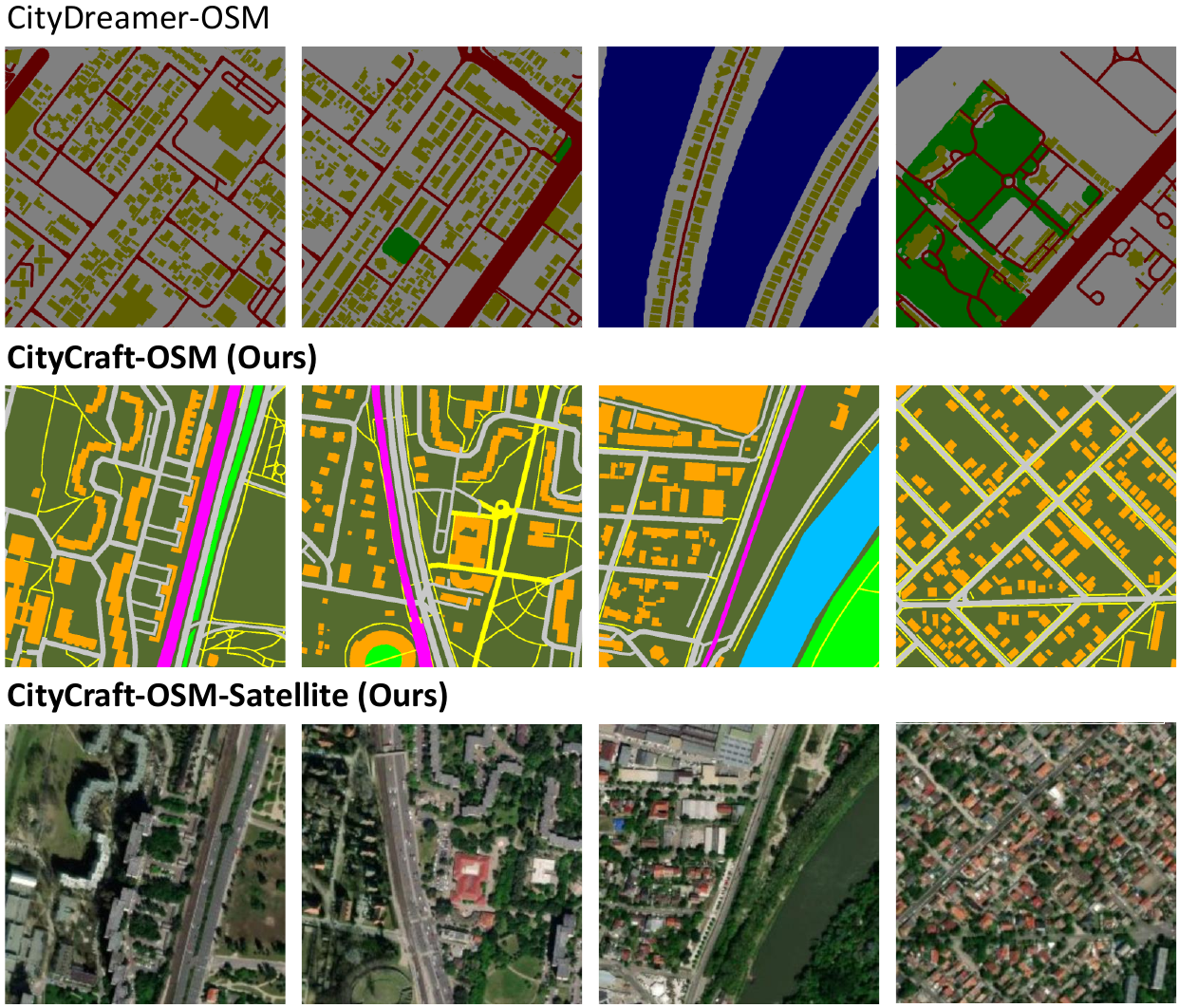}
  \caption{\textbf{Comparison between the CityCraft-OSM and CityDreamer-OSM datasets.} Top row: CityDreamer-OSM~\cite{xie2023citydreamer} visualizations with dense urban details. Middle row: CityCraft-OSM (ours) showcases enhanced clarity and organization in urban mapping. Bottom row: CityCraft-OSM-Satellite (ours) provides real-world satellite imagery, reflecting the accuracy of our city scene representations compared to actual geographical layouts. This comparison highlights our dataset's clarity, usability, and real-world accuracy improvements.
}
  \label{fig:osm_dataset}
\end{figure}

As shown in Figure~\ref{fig:osm_dataset}, we compare the CityCraft-OSM dataset and the CityDreamer-OSM dataset to highlight our dataset's enhancements. The CityCraft-OSM dataset focuses on enhancing the clarity and organization of map data, improving legibility and computational efficiency. Additionally, including real-world satellite imagery in the CityCraft-OSM-Satellite dataset provides a ground-truth comparison, showcasing our dataset's accuracy and practical alignment with real-world urban geographies. These advancements make CityCraft a valuable tool for urban planners, architects, and developers looking to create more realistic and functional urban simulations.

\begin{figure}[t]
  \centering
  \includegraphics[width=\textwidth]{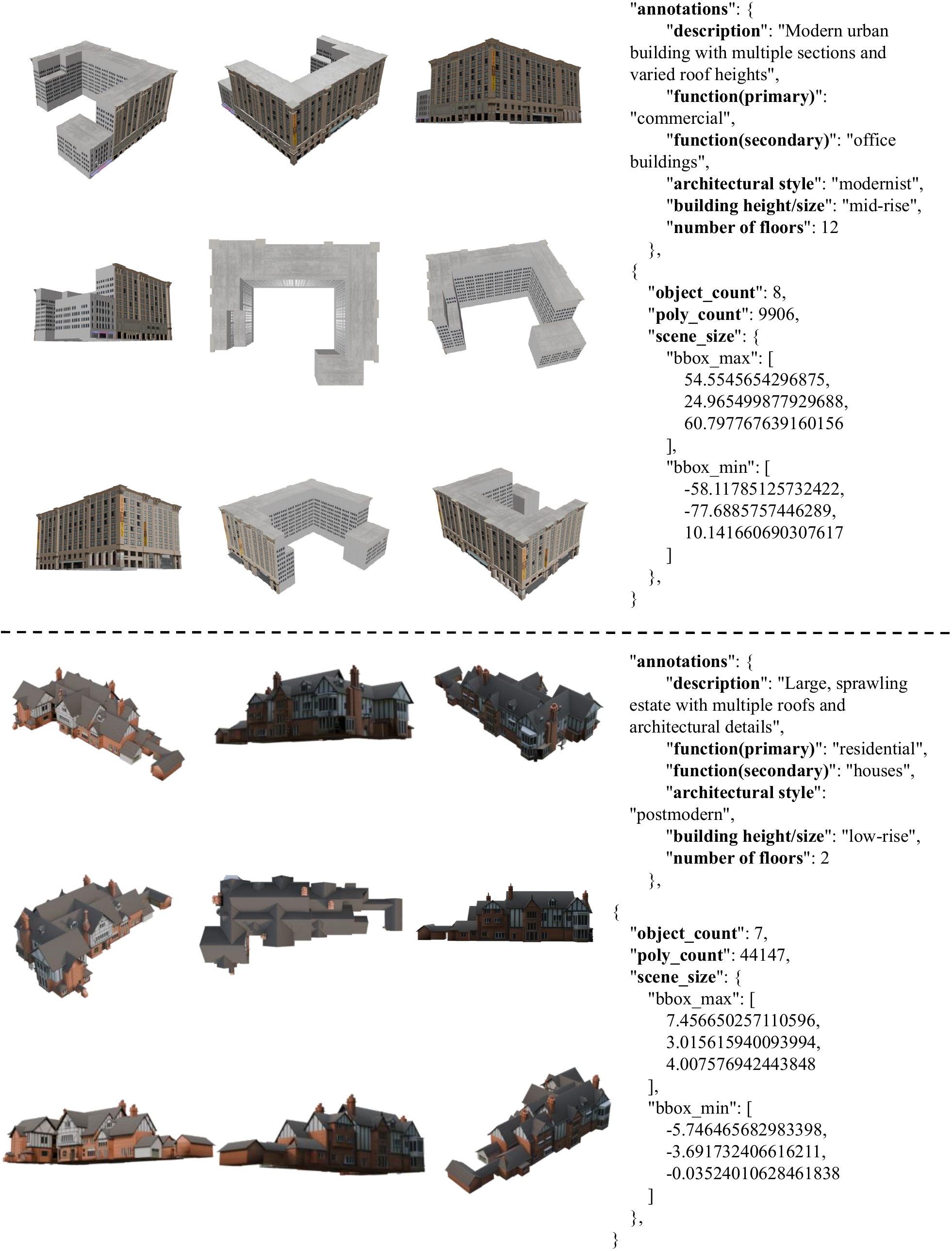}
  \caption{\textbf{3D model examples of architectural assets with corresponding metadata.} The top row features modern-style commercial buildings with varied sections and roof heights, identified as mid-rise (12 floors). The bottom row showcases a large, sprawling residential estate with multiple roofs and intricate details, classified as low-rise (2 floors). Each asset is presented from multiple viewpoints alongside its metadata, including function, architectural style, and object count.}
  \label{fig:asset_demo}
\end{figure}

As shown in Figure~\ref{fig:asset_demo},

\section{User Study Settings}
To better asses the quality of the generated city layouts and the generated city scenes, we conduct user studies and invite 22 participants, including 12 undergraduate students, 8 graduate students, and 2 faculty members. We set the following metrics for evaluating image quality: fidelity, style, consistency, controllability, clarity, sharpness, and overall quality. Each metric has a score from 0 \textasciitilde 10, where a score of 10 is perfect, 8 \textasciitilde 9 is excellent, 6 \textasciitilde 7 is good,  4 \textasciitilde 5 is average, 2 \textasciitilde 3 is poor and 0 \textasciitilde 1 is terrible. In each experiments, we provide the users a set of images generated from different models, and let them give ratings for each model. We compute the final score of each model as the average score of all user's rating, excluding the highest and lowest scores.

\label{sec:user_study_settings}

\section{Demo Visualization}
\begin{figure}
  \centering
  \includegraphics[width=\textwidth]{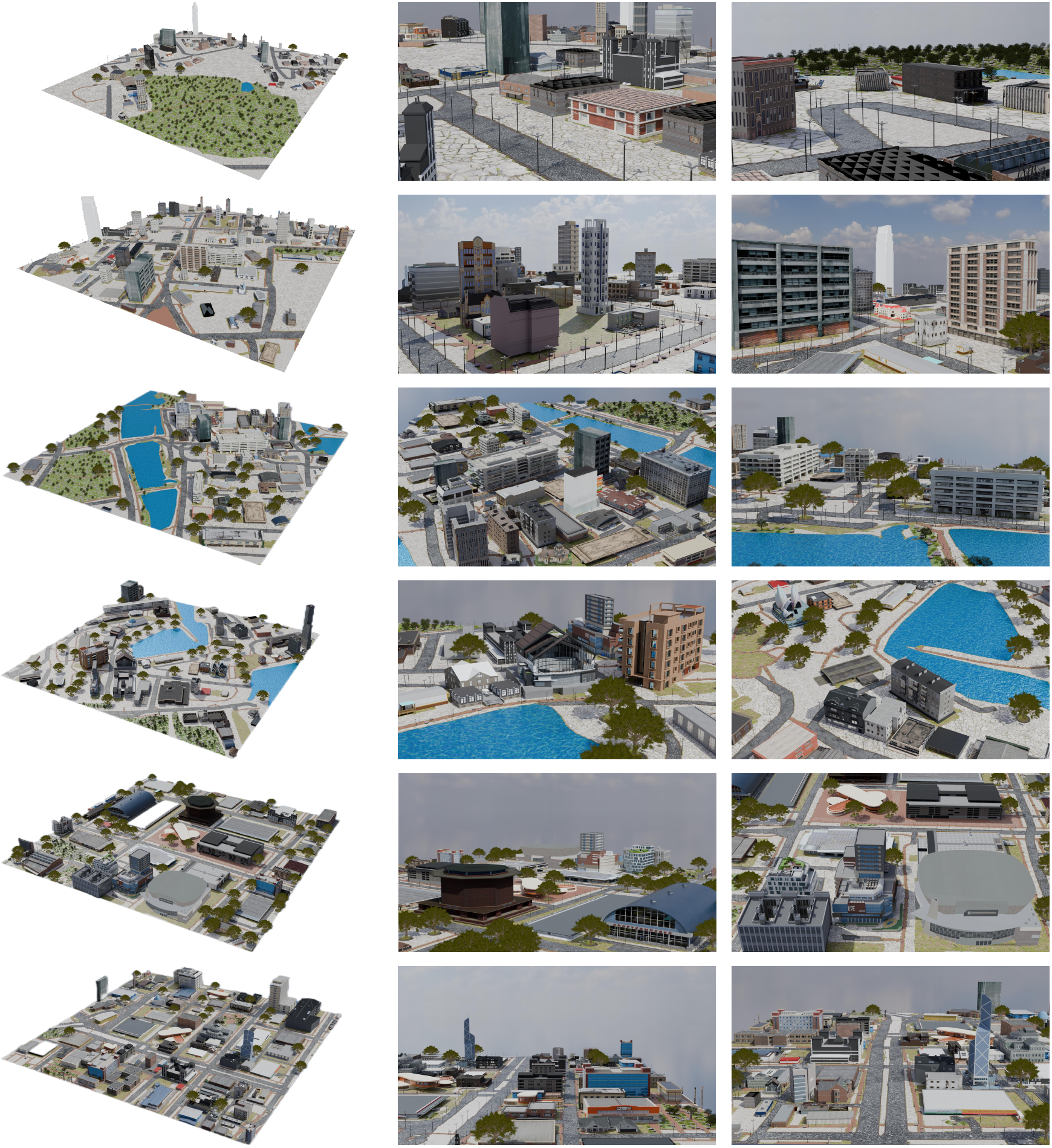}
    \caption{\textbf{Demo visualization} of city scenes generated by CityCraft. It demonstrates the framework's capacity to render complex city environments with high realism. Top: An overall cityscape view showing varied architectural styles and detailed urban planning. Bottom left: A closer view highlighting the intricate modeling of public spaces and individual buildings. Bottom right: A street-level perspective provides a realistic visualization of the urban environment, emphasizing texture quality and integrating natural elements like trees and parks.}
    \label{fig:demo_visualization}
\end{figure}

This section presents a series of demonstrative visualizations highlighting the advanced capabilities of CityCraft in generating highly detailed and realistic city scenes. As shown in Figure~\ref{fig:demo_visualization}, these visualizations serve not only as a testament to the technological prowess of CityCraft but also provide insight into its practical applications in urban planning and architecture.

\subsection{Comprehensive Cityscape Rendering}
The top image offers a panoramic view of the urban landscape generated by CityCraft, showcasing a diverse array of building architectures and meticulously planned urban layouts. This view illustrates how CityCraft handles complex spatial relationships and urban density precisely, facilitating a realistic portrayal of large-scale urban environments.

\subsection{Detailed Environmental Interactions}
The bottom left image focuses on specific areas within the city, such as public squares, parks, and individual buildings. This visualization emphasizes the detailed textural and material qualities that CityCraft can achieve, from the surface textures of roads and pathways to the varied facades of buildings. The attention to detail in these components underscores the model's utility in simulating real-world conditions and enhancing urban design's visual and functional aspects.

\subsection{Street-Level Perspectives}
The bottom right image provides a street-level perspective that brings the viewer into the urban environment created by CityCraft. This perspective is crucial for evaluating the human-scale aspects of urban design, such as walkability, the integration of greenery, and the interaction between different urban elements. It allows planners and architects to assess how well the virtual environment aligns with human-centered design principles.

These demonstrations collectively display the versatility and depth of CityCraft's rendering capabilities, offering stakeholders a powerful tool for visualizing and planning future urban developments with unprecedented detail and realism.

%% file: pseudocode/method.tex
\begin{algorithm}[t]
\caption{CityGen-2 Urban Scene Generation Process $\mathcal{F}$}
\label{alg:citycraft}
\begin{algorithmic}[1]
\STATE \textbf{Input:} User input $X = \{\mathbf{Text}, \mathbf{Ratio}\}$

\STATE \textbf{Output:} Generated urban scene $Y$

\STATE \textbf{Initialize:} 
\STATE $\mathbf{Condition} \leftarrow \mathcal{E}(X)$ \COMMENT{Encode user input into condition vector}

\STATE \textbf{Step 1: Generate Initial Layout}
\STATE $\epsilon \sim \mathcal{N}(0, I)$ \COMMENT{Generate initial noise vector}
\STATE $\mathbf{Layout} \leftarrow \mathcal{G}(\epsilon, \mathbf{Condition})$ \COMMENT{Generate layout using noise and condition}

\STATE \textbf{Step 2: Refine Layout into Urban Plan}
\STATE $\mathbf{Plan} \leftarrow \mathcal{P}(\mathbf{Layout})$ \COMMENT{Refine layout to create detailed plan}

\STATE \textbf{Step 3: Asset Selection and Scene Adaptation}
\STATE $Y \leftarrow \mathcal{A}dapt(\mathbf{Plan}, \mathbf{Layout})$ \COMMENT{Select and place assets according to the plan and layout}

\RETURN $Y$ \COMMENT{Return the final rendered 3D urban scene}
\end{algorithmic}
\end{algorithm}